\documentclass[twocolumn,letterpaper]{IEEEAerospaceCLS}  

\usepackage[]{graphicx}    
\usepackage{amsfonts}
\usepackage{eqnarray,amsmath}
\usepackage{lipsum}
\newcommand{\ignore}[1]{}  

\usepackage{xcolor}

\graphicspath{{../Figures/}}

\begin{document}
\title{Multi-scale Geometric Summaries \\ for Similarity-based Sensor Fusion}

\author{%
Christopher J. Tralie\\
Department of Mathematics \\
Duke University\\
ctralie@alumni.princeton.edu
\and
Paul Bendich\\
Department of Mathematics \\
Duke University\\
Geometric Data Analytics, Inc.\\
bendich@math.duke.edu
\and
John Harer\\
Department of Mathematics\\
Duke University\\
Geometric Data Analytics, Inc.\\
harer@math.duke.edu
\thanks{\footnotesize 978-1-5386-2014-4/18/$\$31.00$ \copyright2018 IEEE}
}

\maketitle

\thispagestyle{plain}
\pagestyle{plain}

\begin{abstract}
In this work, we address fusion of heterogeneous sensor data using wavelet-based summaries of fused self-similarity information from each sensor.
The technique we develop is quite general, does not require domain specific knowledge or physical models, and requires no training. Nonetheless, it can perform surprisingly well at the general task of differentiating classes of time-ordered behavior sequences which are sensed by more than one modality. As a demonstration of our capabilities in the audio to video context, we focus on the differentiation of speech sequences.

Data from two or more modalities first are represented using self-similarity matrices(SSMs)  corresponding to time-ordered point clouds in feature spaces of each of these data sources; we note that these feature spaces can be of entirely different scale and dimensionality.

A fused similarity template is then derived from the modality-specific SSMs using a technique called similarity network fusion (SNF). We investigate pipelines using SNF as both an upstream (feature-level) and a downstream (ranking-level) fusion technique. Multiscale geometric features of this template are then extracted using a recently-developed technique called the scattering transform, and these features are then used to differentiate speech sequences. This method outperforms unsupervised techniques which operate directly on the raw data, and it also outperforms stovepiped methods which operate on SSMs separately derived from the distinct modalities. The benefits of this method become even more apparent as the simulated peak signal to noise ratio decreases.

\end{abstract}

\tableofcontents

\section{Introduction}

This paper suggests a very general and entirely unsupervised heterogeneous sensor fusion pipeline that consists of three geometry-based techniques.
We demonstrate this pipeline on a well-studied digit recognition problem (Section \ref{sec:RP}) in audio-visual fusion, but we claim
that this will be broadly useful in any fusion problem where each sensor produces a time series of arbitrary dimension.

Time series from each sensor are summarized using a two-dimensional construct called a \emph{self-similarity matrix} (SSM, Section \ref{sec:SSM}).
A time series $f(t_1), \ldots, f(t_n)$ of arbitrary dimension produces an SSM whose $(i,j)$ entry encodes the similarity between $f(t_i)$ and $f(t_j)$.
The two time series considered in this work (Section \ref{sec:DP}) correspond to lip pixel frames which are resized to 25x25 pixels from the video sensor (treated as a 625 dimensional Euclidean space) and a sequence of MFCC coefficients \cite{bogert1963quefrency} of dimension 20 from the audio sensor, but we emphasize that the techniques are general enough to handle
time series of any form from any sensor modalities.
This two-dimensional representation of time series from disparate sensor streams enables useful comparisons, facilitating many applications of SSMs, including to cover song retrieval via MFCC features \cite{tralie2015coversongstimbral}, cross-modal comparison \cite{tralie2018CrossModal} and action recognition \cite{junejo2011view}. 

Upstream fusion of SSMs arising from two or more modalities can then be performed (Section \ref{sec:FO}) using \emph{similarity network fusion} (SNF, \cite{wang2012unsupervised,wang2014similarity}),
a random-walk-based technique which is designed to create an SSM which combines the strengths of the individual matrices.
SNF has been applied to different pre-processed modalities arising from musical audio \cite{tralie2017cover} and to improving object level comparisons between 2D shapes\cite{wang2012unsupervised}, cancer phenotypes \cite{wang2014similarity,chen2018ci}, and image collections \cite{chen2018ci}, but to our knowledge, this is the first paper that does so across audio and video modalities for individual objects.
It is also possible (Section \ref{sec:DSNF}) to apply SNF to SSMs defined on the \emph{object level} rather than at the feature level, leading to a downstream fusion technique.

Summary features are then extracted from the fused SSM using the \emph{scattering transform} (Section \ref{sec:ST}). 
This wavelet-based technique (\cite{mallat2012group,Bruna2013}) uses the architecture of a convolution neural network, but without any supervision, to extract a hierarchy of geometric features from
images (including fused and/or non-fused SSMs) in a manner that is provably robust to deformations and preserves multi-scale frequency information.
The scattering transform has been applied in several venues, for example to texture discrimination \cite{sifre2013rotation}.

We advocate combining these techniques into an extremely general pipeline (Figure \ref{fig:Pipeline}) that is entirely unsupervised (i.e, no training data required), does not require domain-specific models, and can handle pre-processed input of any form. 
The benefits of this pipeline are demonstrated via several experiments with increasing levels of simulated noise (Section \ref{sec:ER}), and we also explore different combinations of the techniques (e.g, using the scattering transform directly on the modality-specific SSMs rather than fusing).

There have been countless papers (\cite{Atrey2010} is a good survey) advocating different approaches to audio-visual fusion,
and some which tackle the specific problem of digit recognition.
The vast majority of these techniques are supervised, requiring labeled examples in order to build a model.
For instance, in a widely-used and excellent method \cite{Sargin2007CCA} which combines Canonical Correlation Analysis (CCA) with Hidden Markov Models (HMMs), the CCA subspace in which the modalities are most correlated needs to be learned on a training set before applying the method to new examples, and a different HMM must also be trained on each class. Many of the most recent and very successful methods
\cite{Lopez2018DeepLips} are based on deep learning and thus require very large numbers of labeled examples.
Our proposed approach, by contrast, is completely unsupervised; that is, we rely on labeled examples only to \emph{evaluate} our models, rather than using them in \emph{training} to generate the features used by our models.

\section*{Acknowledgments}
All three authors were partially supported by the Air Force Office of Scientific Research under grant AFOSR FA8750-17-C-0054, as well as
the National Science Foundation under the BIGDATA grant, \# DMS 144749.
We are very grateful to Dr. Erik Blasch (AFOSR) and Drs. Peter Zulch and Jeffrey Hudack (AFRL) for motivating discussions.

\section{Recognition Problem}
\label{sec:RP}

\begin{figure}
    \centering
    \includegraphics[width=\columnwidth]{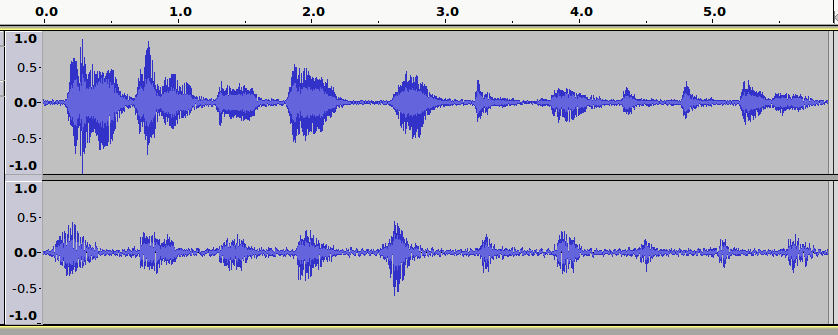}
    \caption{Even after uniformly scaling sequences from two speakers, the raw audio signals do not align perfectly.}
    \label{fig:DigitsUniformScale}
\end{figure}

We demonstrate our methods via a recently-released but already well-studied audiovisual database \cite{Anina2015Oulu}, which
has speakers making utterances while recorded simultaneously by several cameras and by audio.
We focus on the ``Digits'' portion of the database, which contains $51$ distinct speakers uttering digit strings of length ten (e.g, $1 7 3 5 1 6 2 6 6 7$).
There are ten distinct strings, with each one uttered three distinct times by each speaker; hence there are $51 \cdot 3 = 153$ examples of each string.

This dataset is particularly interesting because the two modalities capture different aspects of speech (e.g, 'p' sounds cannot be distinguished from 'b' sounds by video). There are also no obvious correlations between the two modalities, unlike that between, say, nadir-measured speed and horizontally-measured doppler as in some of our earlier work \cite{tralie2018CrossModal}.
Furthermore, there is quite a bit of variation for a single digit string when uttered by different speakers (see Figure \ref{fig:DigitsUniformScale}), and indeed sometimes even by the same speaker.

\subsection{Evaluation Methods}
\label{sec:EM}

Here we rigorously define our problem, as well as the metrics used to evaluate success of the different proposed pipelines.
Let $\mathcal{D}$ be the collection of all utterances of all 10-digit strings, decomposed as the disjoint union of sets $D_i,$ each of which contains the $153$ distinct utterances of the $i$th string, for a total of $1530$ strings in the entire database.
Given any notion of distance $\mu$ on $\mathcal{D}$ and a specific string $s$, the strings in $\mathcal{D}$ can be ranked in increasing order of distance from $s$.
The goal is to create a $\mu$ where, for each such $s$, if $s \in D_i$, then the other strings in $D_i$ will be ranked higher than the strings not in $D_i$.
Success in achieving this goal for a specific $s$ is measured by a {\em precision recall curve}. More precisely, $s$ is pulled out of the database, and the remaining $1529$ strings are put in ranked order by $\mu$.  The ranks of the remaining $152$ digits in the digit class of $s$ are used to construct the precision recall curve as follows; there are $152$ points in the graph, with ``recall'' $r_i$ on the x-axis equal to $r_1 = 1/152$, $r_2 = 2/152$, ..., $r_{152} = 1.0$, and the corresponding ``precision'' values $p_i$ on the y-axis equal to the proportion of items in the correct class to all items we've gone through by the time we reach the $i^{\text{th}}$ correct item.  For instance, if the tenth item occurs at rank 100 in the list, then the recall $r_{10} = 10/152$, and the precision is $p_{10} = 10/100 = 0.1$.  The area under the precision-recall curve is referred to as the {\em mean average precision (MAP)}.  A MAP of $1.0$ is considered perfect performance, while MAP values closer to zero indicate bad performance.  To report a summary statistic for all of $\mu$, we average precision recall curves for all strings in the database $\mathcal{D}$, and compute the corresponding MAP.

Figure \ref{fig:PRresults} displays precision-recall curves and MAPs for the different ranking mechanisms explored in this paper (which we will explain in subsequent sections), and Figure~\ref{fig:NoiseImpactByDigitSequence} explores the impact of noise, which we will explain more in Section~\ref{sec:ER}.  We also compute precision recall curves and MAPs for two other ways of partitioning the data into distinct classes.  First, we decompose $\mathcal{D}$ into $51$ different classes of size $30$, where each class contains all sequences uttered by a particular speaker (i.e. each PR curve has 29 points).  Figure~\ref{fig:NoiseImpactBySpeaker} shows MAPs for this case over different noise levels.  Finally, we explore a finer grained sorting into the intersection of speaker and digit class, for a total of $510$ different classes of size 3 (i.e. each PR curve has 3 points), and figure~\ref{fig:NoiseImpactByDigitSequenceAndSpeaker} shows MAPs for this case.

%
%
%

\begin{figure*}
    \centering
    \includegraphics[width=\textwidth]{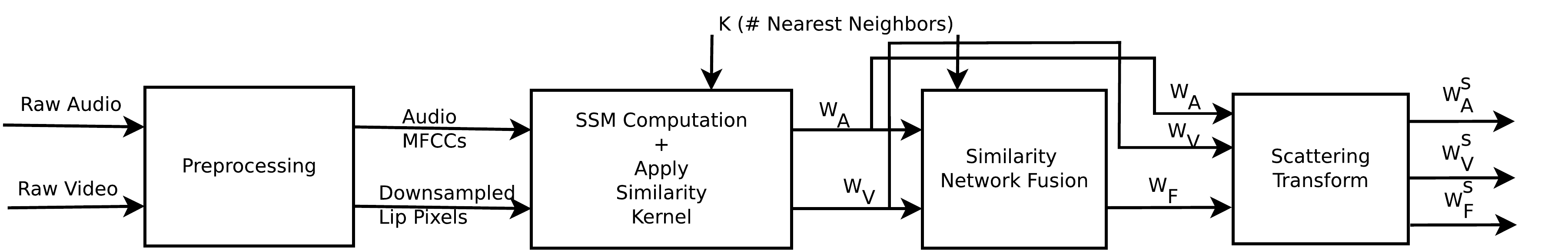}
    \caption{The pipeline for our upstream fusion and data summaries (Sections ~\ref{sec:DP} through ~\ref{sec:ST}).  Note that the only input parameter is the nearest neighbors $K$.  So this is a nearly non-parametric, unsupervised model.  This pipeline can also be easily generalized to handle any number of input features of any type.}
    \label{fig:Pipeline}
\end{figure*}

\section{Data Preprocessing}
\label{sec:DP}


We are now ready to explain our pipeline for comparing the 10 digit strings in more detail.  First, we apply some simple preprocessing to the raw audio and video for each string, which is very similar to preprocessing done by the authors of \cite{Sargin2007CCA}.  For the video, we use pre-extracted lip regions for each speaker that are all resized uniformly to 25x25 grayscale pixels, so that size variations in the lips between subjects due to anatomy and camera position are factored out.  Unlike \cite{Sargin2007CCA}, who perform a zigzagged/shrunken Discrete Cosine Transform (as in the JPEG standard) to further clean this up, we simply use the raw pixels.  Each frame in the video then amounts to one sample in a time series which takes values in 625 Euclidean dimensions.  For the audio, we resample it to 22050hz and compute 20 MFCC coefficients \cite{bogert1963quefrency}, using a window size of 4096 and a hop size of 256.  MFCC coefficients can be viewed as a lossy coarse description of the spectrum in a small window of audio.  As the windows jump by an interval equal to the hop size, this then traces out a time series taking values in a 20 dimensional Euclidean space.

\section{Self-Similarity Matrices}
\label{sec:SSM}

This section provides basic background on Self-similarity Matrices (SSMs), which we advocate as a very useful and general tool
for understanding the time evolution of a process in an arbitrary feature space.
An earlier paper \cite{tralie2018CrossModal} of the authors used SSMs in the heterogeneous sensor framework,
and much of this section is adapted from that paper as well as from the dissertation \cite{tralie2017geometric} of the first author.

\begin{figure}
\centering
\includegraphics[width=\columnwidth]{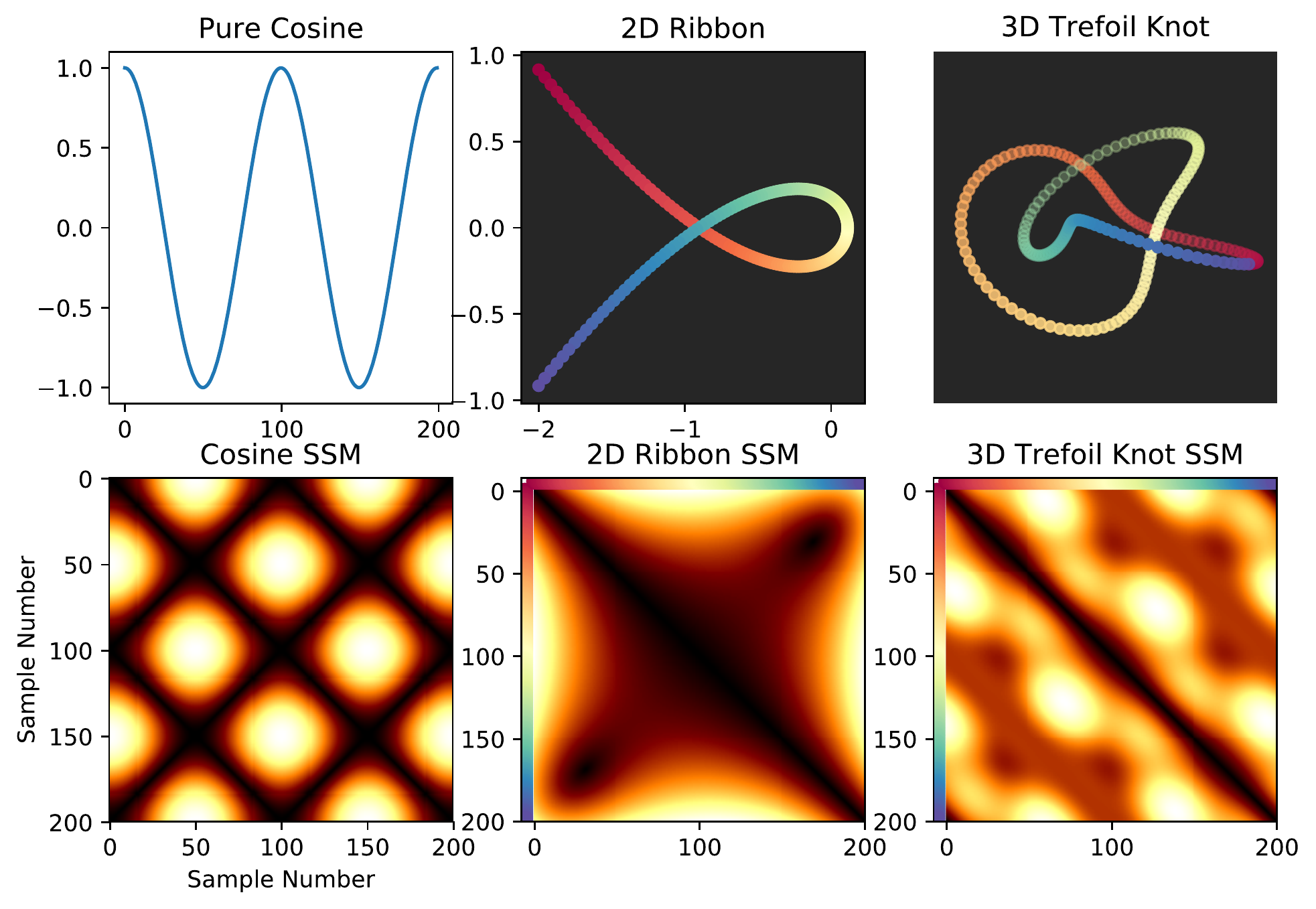}
\caption{Three examples of self-similarity matrices (SSMs) built on three TOPCs: A 1D cosine, a 2D ribbon, and a 3D knot.  Time is on the horizontal axis in the left panel, and is indicated by color in the other two panels.
This was Figure 1 in \protect \cite{tralie2018CrossModal}.}
\label{fig:SSMConcept}
\end{figure}

Suppose that $\gamma: [a,b] \to (M,\rho)$ is a space curve defining a trajectory in some metric space.
Such a curve gives rise to a \emph{self-similarity image} (SSI) $D_{\gamma}: [a,b] \times [a,b] \to \mathbb{R}$ via $D_{\gamma}(x,y) = \rho(\gamma(x), \gamma(y))$
for all pairs of time points $x,y$.
If we discretize the time domain $a = t_1 < t_2 < \ldots < t_N = b$, then we have a time series or \emph{time-ordered point cloud} (TOPC) $X_1, \ldots, X_N \in M$, where $X_i = \gamma(t_i)$.
Then the SSI becomes an $N \times N$ \emph{self-similarity matrix} SSM with $D_{ij} = \rho(X_i, X_j)$.
Three notional examples of SSMs appear in Figure \ref{fig:SSMConcept}.

In this paper, we transform each 10-digit string $s$ into SSMs $D_V(s)$ and $D_A(s)$ by employing the preprocessing described in Section \ref{sec:DP} and using standard Euclidean distance in $R^{625}$ and $R^{20}$, respectively.
Note that the entries in an SSM can go from zero to infinity, where small entries indicate close proximity.
To facilitate the probability-based methods in Section \ref{sec:FO}, we transform each SSM as follows.
Given any TOPC point cloud $X_1, \ldots, X_N$ and any metric $\rho$ we define a {\em similarity kernel $W$} as
\begin{equation}
    \label{eq:similaritykernel}
    W_{ij} = \exp \left( -\frac{\rho^2(x_i, x_j)}{\sigma_{ij}} \right)
\end{equation}

This is a similarity measure rather than a distance, as points which are far have a score close to 0, and points which are close have a score closer to $1$.  Often, $\sigma_{ij}$ is set to a constant, but a smarter choice is to autotune it based on the average distance to the nearest neighbors of $x_i$ and $x_j$.  In particular, we can follow  \cite{wang2014similarity} and set:
\begin{eqnarray*}
\sigma^{\kappa}_{ij} & = & \frac{\beta}{3} (\frac{1}{\kappa N} ( \sum_{k \in N^{\kappa}(i)} \rho(x_i, x_k) ) \\
& & + ( \frac{1}{\kappa N} \sum_{k \in N^{\kappa}(j)} \rho(x_j, x_k) ) +  \rho(x_i, x_j) ), 
\end{eqnarray*}
where $\kappa$ is the proportion of nearest neighbors taken (we use $\kappa = 0.1$ in this work), $N^{\kappa}(i)$ refers to the $\kappa N$ nearest neighbors of $x_i$, and $\beta$ is a parameter that can be tweaked (usually in the range $[0.3, 0.8]$). Of course, kernels other than the Gaussian are possible, but we find this works well in practice for our applications.

\begin{figure}
    \centering
    \includegraphics[width=\columnwidth]{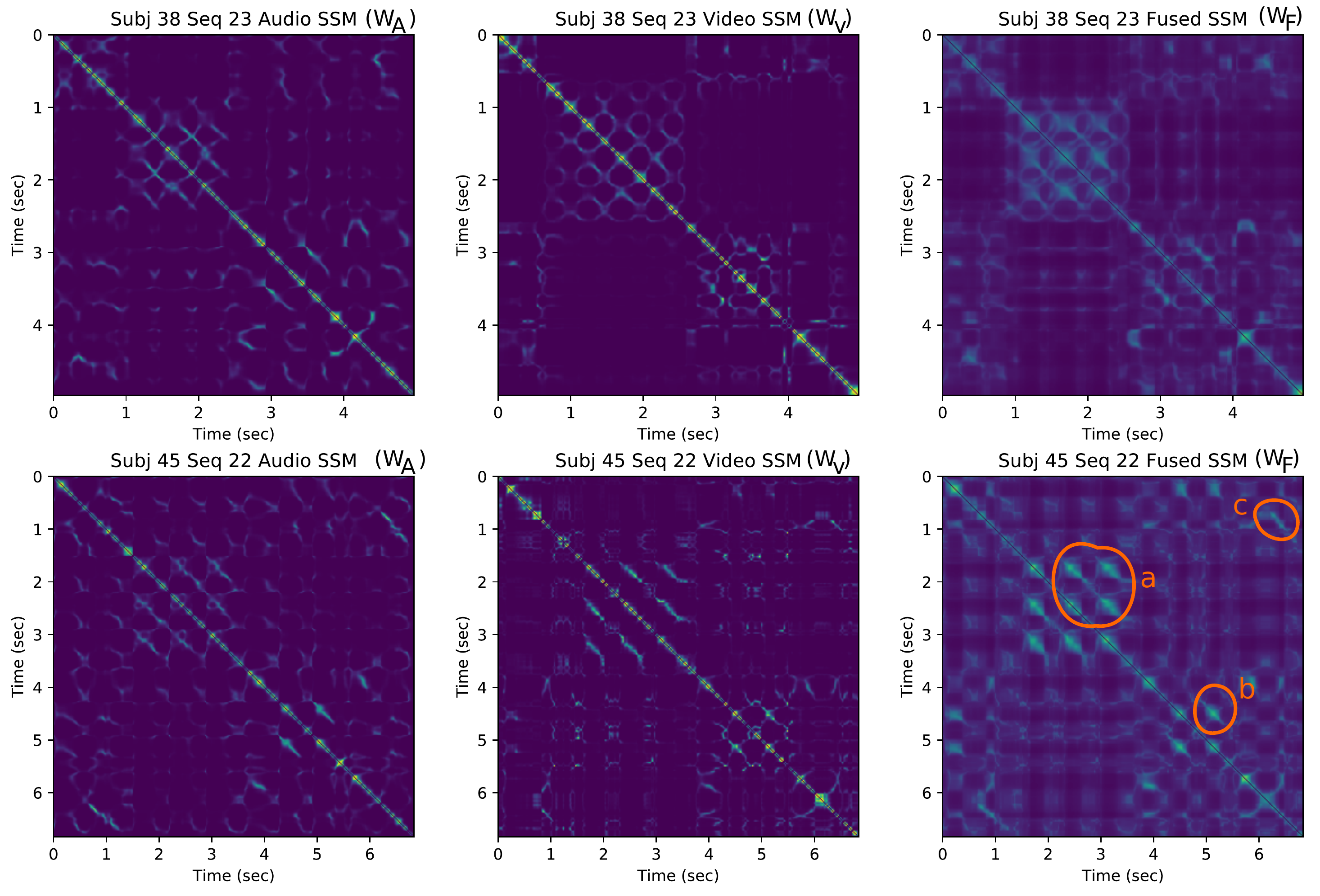}
    \caption{SSMs for the digit sequence ``9 7 4 4 4 3 5 5 8 7'' spoken by two different speakers.  The SSMs for different speakers are on each row, the first column shows SSMs for audio, the second column shows SSMs for video, and the third column shows the fused SSMs.  Bright means similar and dark means dissimilar.  The repetitions of the two 4s are circled in region a in the bottom speaker, the repetitions of the 5 digit is circled in region b, and the repetition of the digit "7" is circled in region c.  These structures are visible in both speakers.}
    \label{fig:SSMsExample}
\end{figure}

\begin{figure}
    \centering
    \includegraphics[width=\columnwidth]{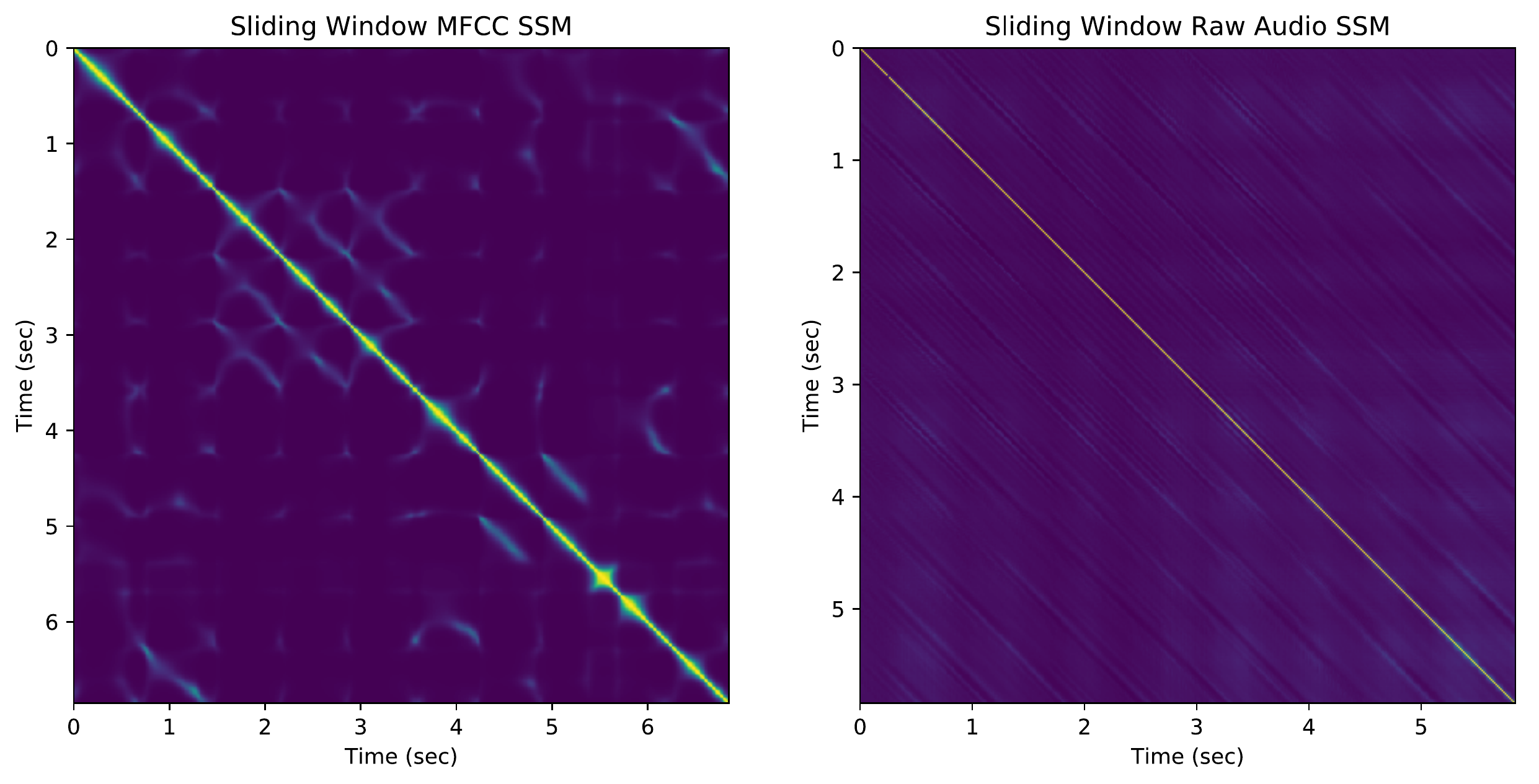}
    \caption{SSMs for audio on the digit sequence ``9 7 4 4 4 3 5 5 8 7'' as summarized by MFCC and as summarized by raw samples.  Structure, such as the repeating 4s, 5s, and 7s, as in Figure~\ref{fig:SSMsExample}), is not visible in the raw samples.}
    \label{fig:RawAudioVsMFCC}
\end{figure}

Applying this procedure to $D_A(s)$ and $D_V(s)$ results in new SSMS, $W_A(s)$ and $W_V(s)$.
Examples of each type appear in the first two columns of Figure \ref{fig:SSMsExample}.
One can of course also compute an SSM directly from the raw audio data, treated as a simple one-dimensional time series.
Figure \ref{fig:RawAudioVsMFCC} compares such an SSM on the right with the MFCC-aided SSM (left), and indicates that the latter picks up on more meaningful structure.

\section{Upstream Similarity Network Fusion}
\label{sec:FO}

Sections \ref{sec:DP} and \ref{sec:SSM} describe how to transform a digit string into two SSMs, with $W_A$ derived from audio and $W_V$ derived from video.
In general, one is often faced with the situation where one has two or more similarity measures defined on the same finite ordered set.
The technique of \emph{similarity network fusion} (SNF, \cite{wang2012unsupervised,wang2014similarity}) takes the SSMs of these similarity measures and outputs
a single fused SSM which is meant to leverage the strengths of each individual SSM.

\begin{figure}
    \centering
    \includegraphics[width=\columnwidth]{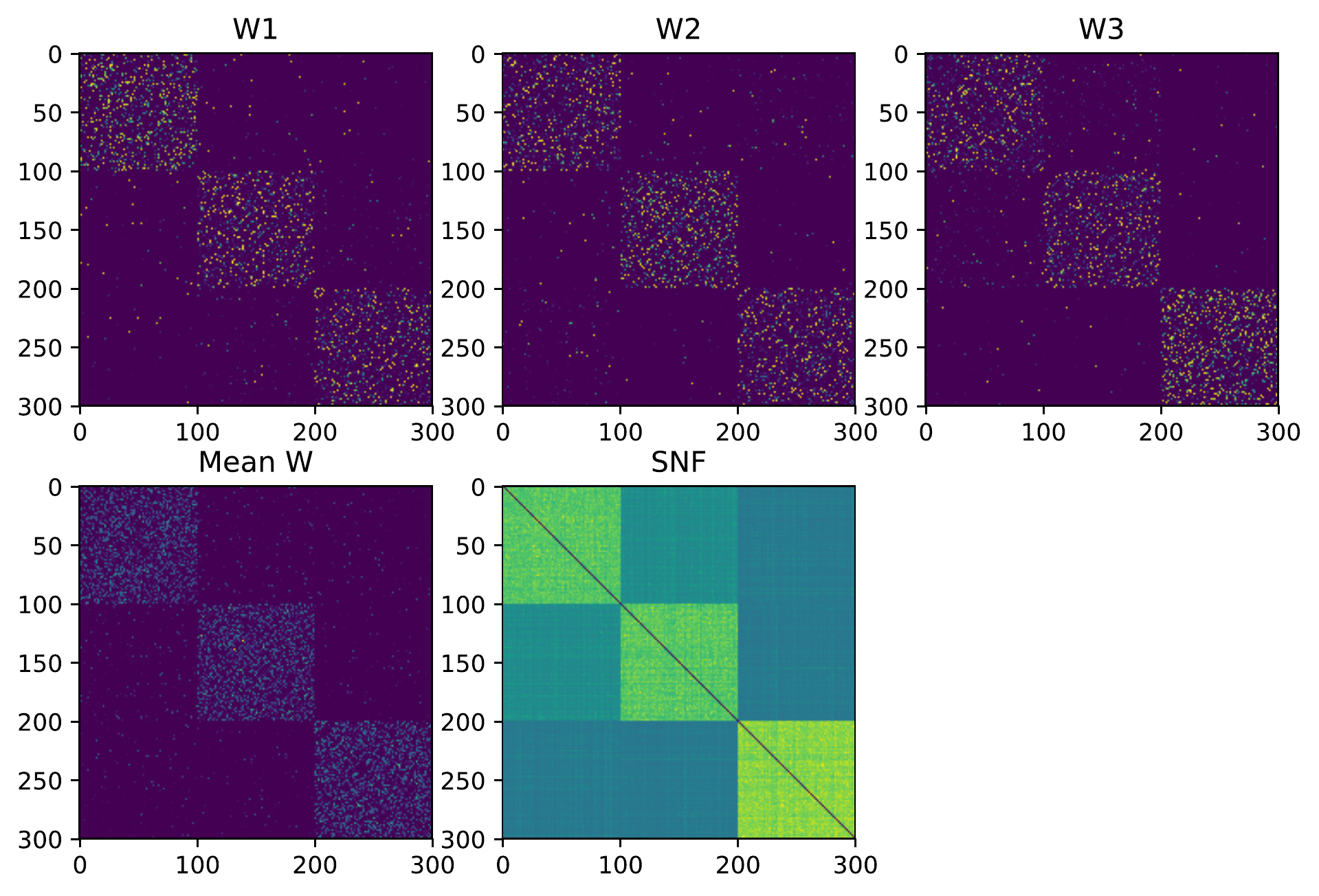}
    \caption{A notional example of similarity network fusion from three different (top row) noisy measurements of 3 simulated clusters.  Simple averaging (bottom row, left) of the similarities is far inferior to similarity network fusion (bottom row, middle).
    }
    \label{fig:SNFConcept}
\end{figure}

A notional example appears in Figure \ref{fig:SNFConcept}.
We imagine that we have a TOPC consisting of three distinct clusters with $100$ points each, and we noisily sample these clusters three separate times.
Each time we do so, the $\ell_2$ distance in the plane gives a similarity measure.
The SSMs corresponding to these three distinct similarity measures appear in the top row of the figure.
While each measure is likely to see a pair of points belonging to a single cluster as more similar to each other (the yellow speckles along the diagonal) than a pair of points from distinct clusters, this is certainly not always the case (the many blue gaps among the yellow speckles).
A similar critique holds for the simple average of the three SSMs.
On the other hand, solid cluster membership is much more apparent in the SSM produced by the SNF algorithm.
The third column of Figure~\ref{fig:SSMsExample} shows fused SSMs resulting from audio and video SSMs in the digits dataset.

We now give some technical details of SNF, referring the reader to \cite{wang2012unsupervised,wang2014similarity} for a fuller description.  Given $M$ different $N \times N$ similarity matrices $W_1, W_2, ..., W_M$ (as in Equation~\ref{eq:similaritykernel}), we normalize them into corresponding matrices $P_1, P_2, ..., P_M$ as follows

\begin{equation}
P_m(i, j) = \left\{  \begin{array}{cc} \frac{W_m(i, j)}{2 \sum_{k \neq i} W_m(i, k)} & j \neq i \\ 1/2 & j = i \end{array} \right\}
\end{equation}

Since each row now sums to 1, each $P_m$ can be thought of as a transition probability matrix on a graph with $N$ vertices.  In addition to the $P$ matrices, we also define an associated ``masked'' transition probability matrix $S_m$ as follows

\begin{equation}
S_m(i, j) = \left\{  \begin{array}{cc} \frac{W_m(i, j)}{2 \sum_{k \in N_i} W_m(i, k)} & j \in N_i \\ 0 & \text{otherwise} \end{array} \right\}
\end{equation}

where $N_i$ is the $\kappa N$ nearest neighbors of point $i$ (those $j$s with the $\kappa N$ largest similarity values in the $i^{\text{th}}$ row of $W^{m}$, excluding $i$).  This is similar to $P$, except that the probabilities outside of the nearest neighborhoods are set to zero, and the remaining entries are reweighted to sum to 1.  Given these sets of matrices, SNF updates the $P$ matrices in an iterative fashion, while holding the $S$ matrices fixed.  Referring to $P_m^t$ as the $P$ matrix for the $m^{\text{th}}$ feature after $t$ iterations, where $P_m^{0} = P_m$, the recursive update rule can be written as the following matrix multiplication

\begin{equation}
\label{eq:SNFiter}
P_m^{t} = S_m \times \left( \frac{\sum_{k \neq m} P_k^{t-1}}{M-1}  \right) \times (S_m)^{\dag}
\end{equation}

Intuitively, this can be thought of as one step of a random walk through neighborhoods determined by $W_m$ using transition probabilities from all of the other $W$s.  At the end of some specified number of iterations $T$ (we use $T = 20$), the final similarity matrix can be obtained as 

\begin{equation}
\frac{1}{M} \sum_{m = 1}^M P_m^T
\end{equation}

Note that for our pipeline so far, we only have $M = 2$ ($W_A$ for audio and $W_V$ for video), though in Section~\ref{sec:DSNF}, we will show a case where $M = 3$.

Due to the matrix multiplication in Equation~\ref{eq:SNFiter}, the time complexity of this algorithm is $O(N^3)$ for an $N \times N$ similarity matrix, though this can be mitigated with a sparse matrix for $S$.  Regardless, this is not a computational bottleneck for the short sequences under consideration in this specific application.

\subsection{Uniformly Rescaling SSMs}
Note that SNF requires all of the $W$ matrices to be the same size and in correspondence; that is, in the case of audio/video SSM fusion for a particular digit sequence by a particular speaker, the $i^th$ row of $W_A$ needs to correspond to the same point in time as the $i^th$ row of $W_V$.  Hence, we simply resize $W_A$ and $W_V$ to a common dimension ($256 \times 256$) with image interpolation.  Furthermore, a common dimension for all SSMs allows us to compare across digit sequences which have different lengths in time, which is important since even the same speaker is unlikely to say the same sequence in exactly the same amount of time during different runs.

\section{Scattering Transform}
\label{sec:ST}

\begin{figure}
    \centering
    \includegraphics[width=\columnwidth]{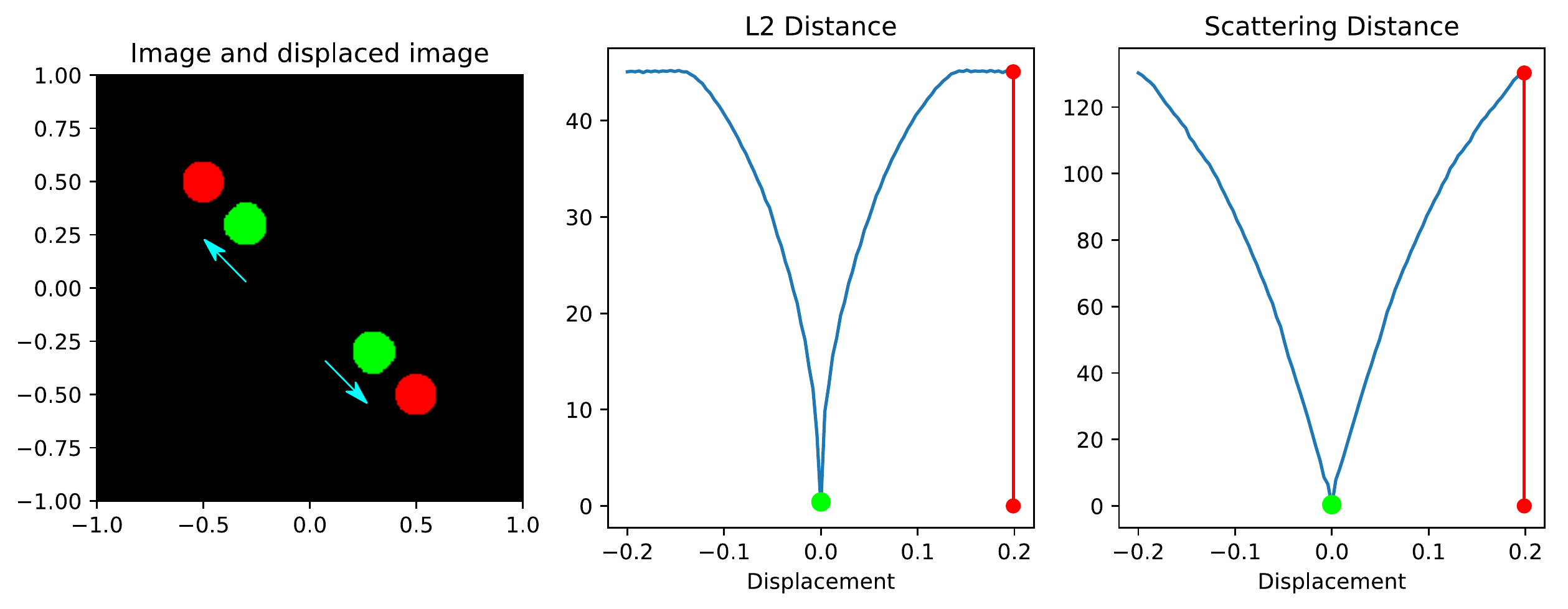}
    \caption{An example of raw L2 versus the scattering transform on an image of two blobs.  The initial blobs are drawn in green and an example of displaced blobs are drawn in red.  The cyan arrow shows the positive direction of displacement.}
    \label{fig:ScatteringConcept}
\end{figure}

Once $W_A$ and $W_V$ have been resized to the same dimension, it is possible to compare instances of each to each other with an L2 norm, otherwise known as a matrix Frobenius norm.  That is, the distance between two $N \times N$ matrices $A$ and $B$ could be defined as

\begin{equation}
    ||A - B||_2 = \sqrt{ \sum_{i=1}^N \sum_{j=1}^N (A_{ij} - B_{ij})^2 }
\end{equation}

For sequences which unfold at the same rate up to a uniform scale, this is a good distance to use.  However, in practice, it is unlikely that two runs of the same digit sequence will line up exactly, even after a uniform rescaling.  Instead, there are often local delays, or ``time warps,'' that will cause them to be out of sync, and which will induce local perturbations in the SSMs\footnote{The authors of \cite{sulam2017dynamical} have made a similar observation for time series in dynamical systems and have used the 1D scattering transform to ameliorate this}.  Unfortunately, when small, high frequency details, such as those in SSMs (Figure~\ref{fig:SSMsExample}), are perturbed slightly spatially, the L2 norm can be unstable.  Figure~\ref{fig:ScatteringConcept} shows a synthetic example of this phenomenon.  In general, one can show this instability mathematically by applying the 2D Fourier transform to the matrix, which is an isometry, and noticing that high frequency bins change dramatically for any small change in spatial position \cite{bruna2013scattering}.  

Since this instability only occurs at fine scales, one could instead try applying an orthogonal 2D wavelet transform \cite{mallat1999wavelet} to the SSMs, which, unlike the Fourier Transform, is spatially localized in a hierarchical fashion.  However, the wavelet transform is also an isometry, so the same instabilities are still present; this amounts to ``bin hopping'' of the wavelet coefficients at the fine scales, which is a problem that also plagues histograms compared with Euclidean distance.  Motivated by this problem, Mallat \cite{mallat2012group} devised a nonlinear alternative to the wavelet transform known as the {\em scattering transform}, which is still hierarchical, but which is stable.  Like a wavelet transform, a particular scattering transform starts with a {\em mother wavelet} $\psi(u, v)$ and its corresponding {\em scaling function}, or lowpass filter $\phi(u, v)$.  However, the scattering transform insists that $\psi(u, v)$ be complex-valued, and in the case of 2D images, it is {\em directional}, so we index it with a direction vector $\psi_{\gamma}(u, v)$, where $\gamma = (\gamma_x, \gamma_y), \gamma_x^2 + \gamma_y^2 = 1$.  Finally, the mother wavelet can be scaled in a dyadic fashion, so that 
\begin{equation}
\psi_{\gamma, s}(u, v) = 2^{-2s}\psi_{\gamma}(u/2^s, v/2^s)
\end{equation}
Then, given an 2D image $I(u, v)$, the level 0 scattering coefficients $S^0(u, v)$ are simply a lowpass filter

\begin{equation}
S^0(u, v) = I \ast \phi (u, v)
\end{equation}

The level 1 scattering coefficients, on the other hand, are computed for each of $L$ directions and each of $J$ scales as follows

\begin{equation}
S^1_{i, j} (u, v) = |I \ast \psi_{\gamma_i, j}(u, v)| \ast \phi(u, v)
\end{equation}

for $i = 1, 2, \hdots L$ and $j = 0, 1, \hdots J-1$, where $|.|$ is the complex modulus (absolute value), and the convolution with the scaling function $\phi$ is followed by downsampling at scale $J$.

Then, the level 2 scattering coefficients are computed for a {\em pair} of wavelets $\psi_{\gamma_i, j}$ and $\psi_{\gamma_k, \ell}$ as follows

\begin{equation}
S^2_{i, j, k, \ell} (u, v) = ||I \ast \psi_{\gamma_i, j} (u, v) | \ast \psi_{\gamma_k, \ell } (u, v)| \ast \phi(u, v)
\end{equation}

for $i = 1, 2, \hdots L$, $j = 0, 1, \hdots, J-1$, $k = 1, 2, \hdots L$, and $\ell = 0, 1, \hdots j-1$. In other words, the level 2 scattering coefficients compute interactions between all pairs of wavelets at different directions and different scales, which allows them to pick up on higher order information not possible with an ordinary Fourier or wavelet transform.  For instance, the scattering coefficient can represent corners as the interaction of two wavelets at different directions, while wavelets and Fourier modes can only represent a gradient along a single direction.  Note that for second order scattering, the scale of the first wavelet $i$ is greater than the scale of the second wavelet $\ell$, since it has been shown that only these combinations yield a non-negligible result \cite{waldspurger2017exponential}.  Said differently, the first wavelet should be at a lower frequency than the second.  Hence, there are $L^2J(J-1)/2$ different combinations of wavelet pair interactions for level 2.

To continue to even higher order scattering coefficients at level $q$, one continues this pattern of convolving with a sequence of wavelets of decreasing scales, always followed by a complex modulus (the nonlinear element), and finished with a lowpass filter, for a total of $L^q {J \choose q}$ wavelet sequences, each also referred to as a ``scattering path.''  Hence, the architecture of the scattering transform is similar to the architecture of a deep convolutional neural network \cite{lecun1995convolutional,lecun2015deep}, although the weights are not learned but instead fixed based on the mother wavelet.  This crucial difference accounts for the stability of the scattering transform versus the empirical instability of deep neural networks that exposes them to adversarial attack in certain contexts \cite{su2017one,Moosavi2016DeepFool}.  It also circumvents the need to learn weights from examples, keeping our pipeline unsupervised.  For a more complete description of the 2D scattering transform in the context of texture classification, please refer to \cite{bruna2011classification}.

In our case, we use complex Morlet (Gabor) wavelets, since they have nearly optimal frequency localization \cite{mallat1999wavelet}.  They take the form of a complex plane wave attenuated by a Gaussian

\begin{equation}
\psi_{\gamma}(u, v) = e^{i \gamma \cdot (u, v) } e^{ -(u^2+v^2)/\sigma^2 }
\end{equation}

with corresponding scaling functions which are the Gaussians $\phi(u, v) = \exp(-(u^2+v^2)/\sigma^2)$.  We resize our images to a $256 \times 256$ resolution (65536 pixels), and we take $J = 4$ scales and $L = 8$ directions equally spaced between $0$ and $\pi$ (since the interval $[\pi, 2 \pi)$ is redundant in directions with $[0, \pi)$).  We then perform two levels of scattering and downsample each scattering path to a $32 \times 32$ resolution after lowpass filtering.  This leaves us with $k = 32^2(1 + 4 \times 8 + (4 \times 3 / 2) \times 8^2 ) = 427008$ scattering coefficients (a roughly 6.5x increase in data size, but at a gain of stability).  The scattering distance can then be taken as the Euclidean distance in this $k$-dimensional Euclidean space, and this is provably stable \cite{mallat2012group}.  As an example, Figure~\ref{fig:ScatteringConcept} shows distances of the scattering transform versus straight L2 distances between images of blobs as the blobs are displaced along a line.  Note that the L2 and scattering distance are both near zero for very small displacements, but the L2 distance maxes out once the blobs no longer overlap at an absolute displacement of 0.1, the radius of the blob.  By contrast, the scattering transform distance continues to be sensitive to changes even beyond the support of the blob, hence demonstrating its stability to small deformations.

\begin{figure}
    \centering
    \includegraphics[width=\columnwidth]{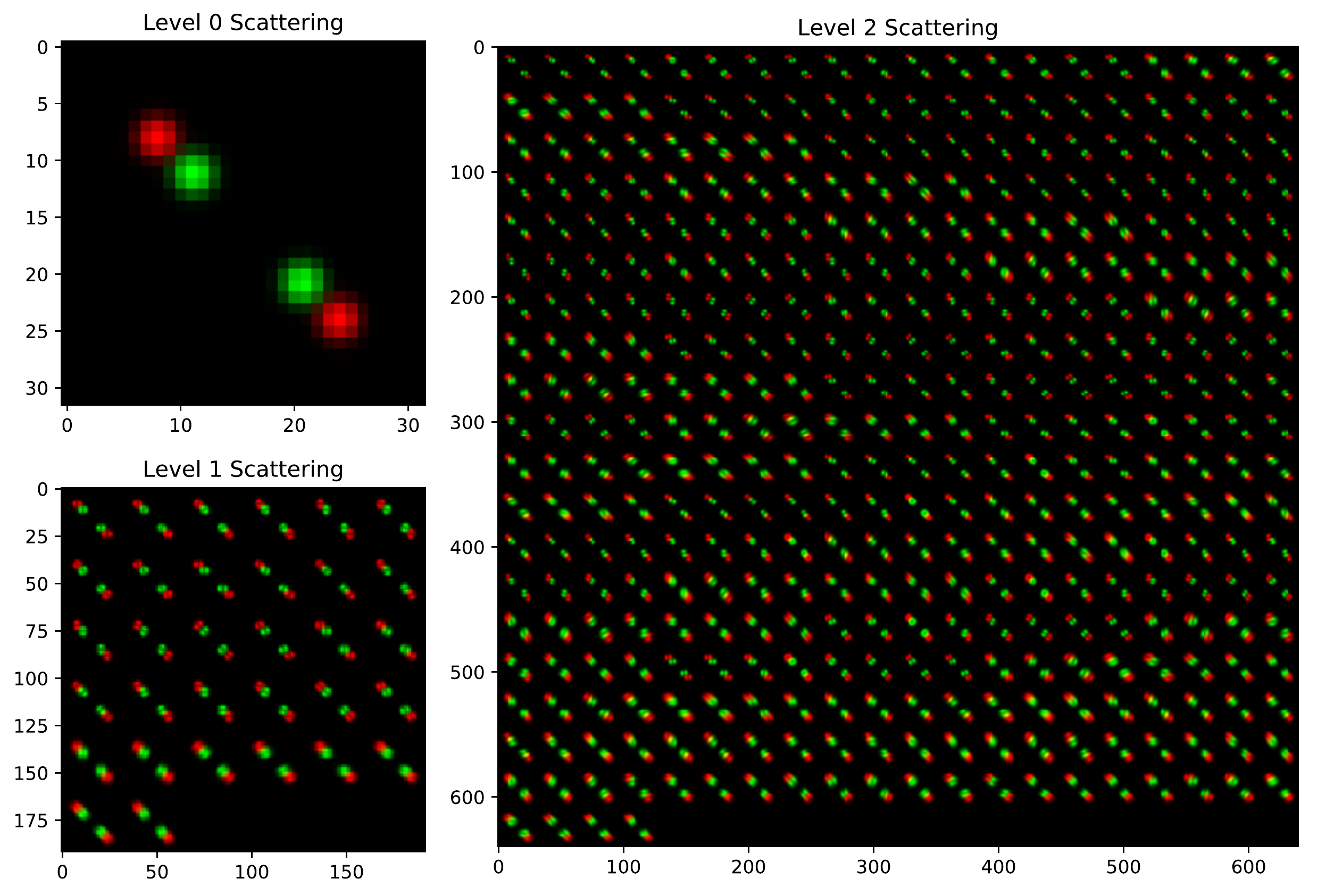}
    \caption{An example of the zeroth, first, and second order scattering coefficients for the blob image and the shifted blob image in Figure~\ref{fig:ScatteringConcept}.  Although they do not overlap in the original images, some of their first and second order scattering coefficients do overlap slightly (the perturbation has been ``smoothed out''), leading to a non-saturated distance.}
    \label{fig:ScatteringExample}
\end{figure}

\begin{figure}
    \centering
    \includegraphics[width=\columnwidth]{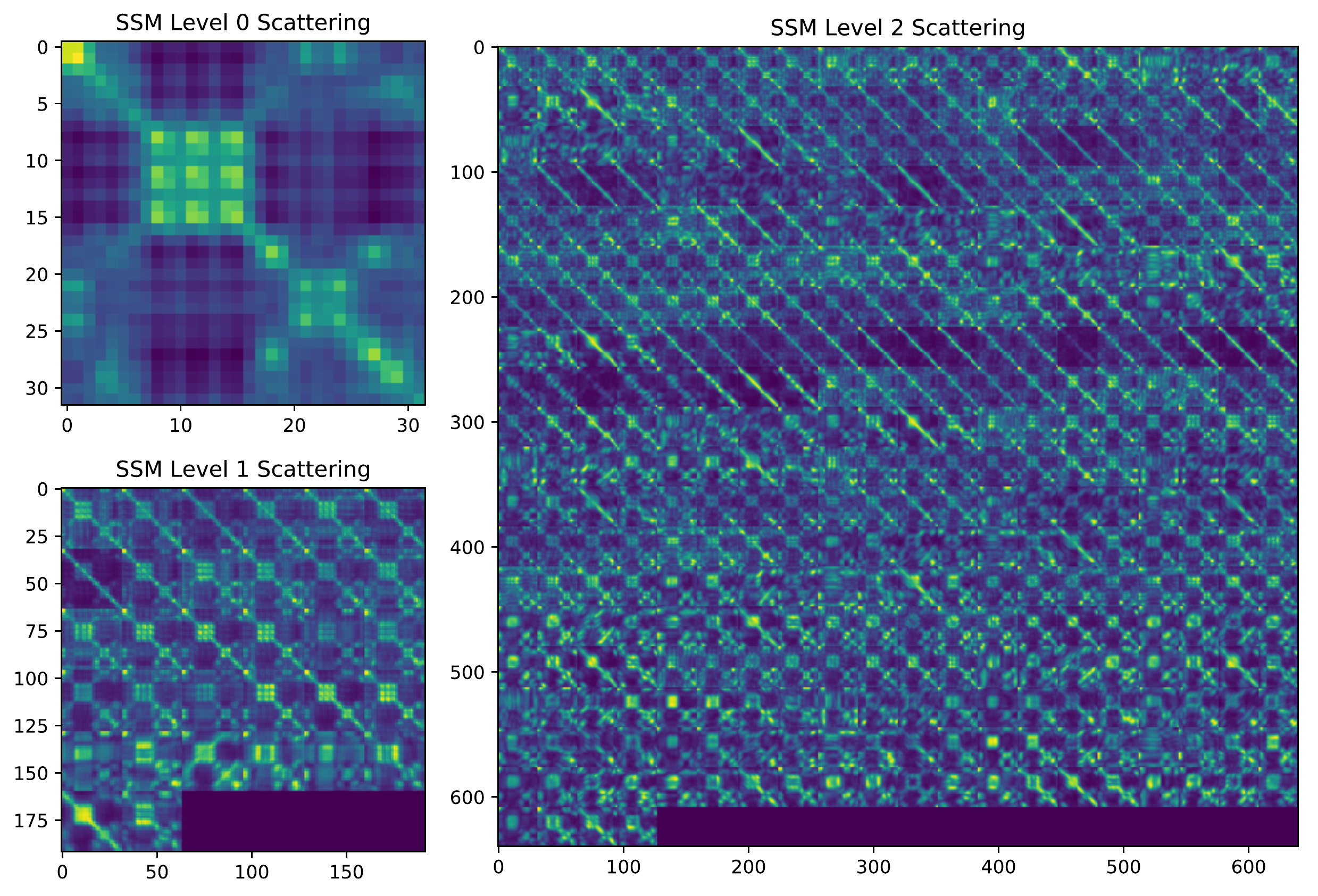}
    \caption{An example of the zeroth, first, and second order scattering coefficients for the fused SSM for subject 45 sequence 22 (bottom right of Figure~\ref{fig:SSMsExample}).}
    \label{fig:SSMScatteringExample}
\end{figure}

In our pipeline, we use the scattering transform exclusively on self-similarity images.  Figure~\ref{fig:SSMScatteringExample} shows an example of the scattering transform for the fused SSM for subject 45 sequence 22 (bottom right of Figure~\ref{fig:SSMsExample}).  Finally, note that, like the 2D wavelet transform, the scattering transform on our $N \times N$ images has time complexity $O(N^2 \log (N))$.  Since $N$ is small for this application, this is not a computational bottleneck for the scattering transform on individual SSMs.

\section{Downstream Similarity Network Fusion}
\label{sec:DSNF}

So far, all of the steps we have described can be performed upstream: that is, before any ranking decisions have been made.  Figure~\ref{fig:Pipeline} shows the flow from step to step.  However, if we have access to $N$ examples of digit sequences already and we perform all pairwise comparisons between them, we can perform one additional, optional step to improve classification, even without using any labels on these sequences.  Let $\mu_1$, $\mu_2$ and $\mu_3$ be three different $N \times N$ matrices measuring all of the pairwise similarities between digits, as measured by some path in our upstream pipeline.  For instance we can use L2 on $W_A$ ($\mu_1$), $W_V$ ($\mu_2$), $W_F$ ($\mu_3$), or we can use L2 on their scattering counterparts $W^S_A$ ($\mu_1$), $W^S_V$ ($\mu_2$), or $W^S_F$ ($\mu_3$).  Note that $\mu_1, \mu_2, \mu_3$ are themselves SSMs, but {\em at the object level}.  Because they are SSMs, we can apply SNF to them to potentially improve the ensuing rankings.  We refer to this as {\em downstream similarity network fusion}, as it happens at the object level after all upstream features have been computed, and we do indeed see that it improves results in some cases in Section~\ref{sec:ER}.  Note that this also works with only two different similarity measures, though we see advantages to using all three at our disposal within L2 or scattering.

\section{Experiments and Results}
\label{sec:ER}

This section puts the geometric tools described in Section \ref{sec:SSM}-\ref{sec:DSNF} together into several different pipelines, and then evaluates these pipelines
on the dataset described in Section \ref{sec:RP}.
As stated above, each pipeline creates a metric $\mu$ on the set of digit strings $\mathcal{D}$, and here they do so by transforming each string into a feature vector in a suitable Euclidean space (either raw SSM entries or scattering coefficients on SSMs) and then using the standard $\ell_2$ metric between pairs of feature vectors in that space.

The method we advocate (Figure \ref{fig:Pipeline}) starts by extracting one audio and one video time series from each string as described in Section \ref{sec:DP}.
These time series are transformed (Section \ref{sec:SSM}) into similarity matrices $W_A(s)$ and $W_V(s)$, which are then fused (Section \ref{sec:FO}) to produce a single SSM $W_F(s)$.
Finally, the scattering transform (Section \ref{sec:ST}) is applied to $W_F(s)$ to produce a sequence of scattering coefficients which form the needed feature vector.
This method is referred to as FusedScatter in the test results below.

We also explore several other pipelines. FusedL2 stops at the third block in Figure \ref{fig:Pipeline}, stacking the entries of the matrix $W_F(s)$ to form the feature vector. 
Similarity, AudioL2 (resp., VideoL2) outputs $W_A(s)$ (resp., $W_V(s)$) as the feature vector.
AudioScatter (resp., VideoScatter) uses only the audio (resp., video) time series, transforms it into $W_A(s)$ (resp., $W_V(s)$), and then directly extracts scattering coefficients without any similarity network fusion.
Finally, we can perform late fusion (Section~\ref{sec:DSNF}) to AudioL2 and VideoL2, or AudioL2, VideoL2, and FusedL2, and similarly for AudioScatter and VideoScatter and AudioScatter, VideoScatter, and FusedScatter.  We call these AVLateFusedL2, AllFusedL2, AVLateFusedScatter, and AllFusedScatter, respectively.
\begin{figure}
    \centering
    \includegraphics[width=\columnwidth]{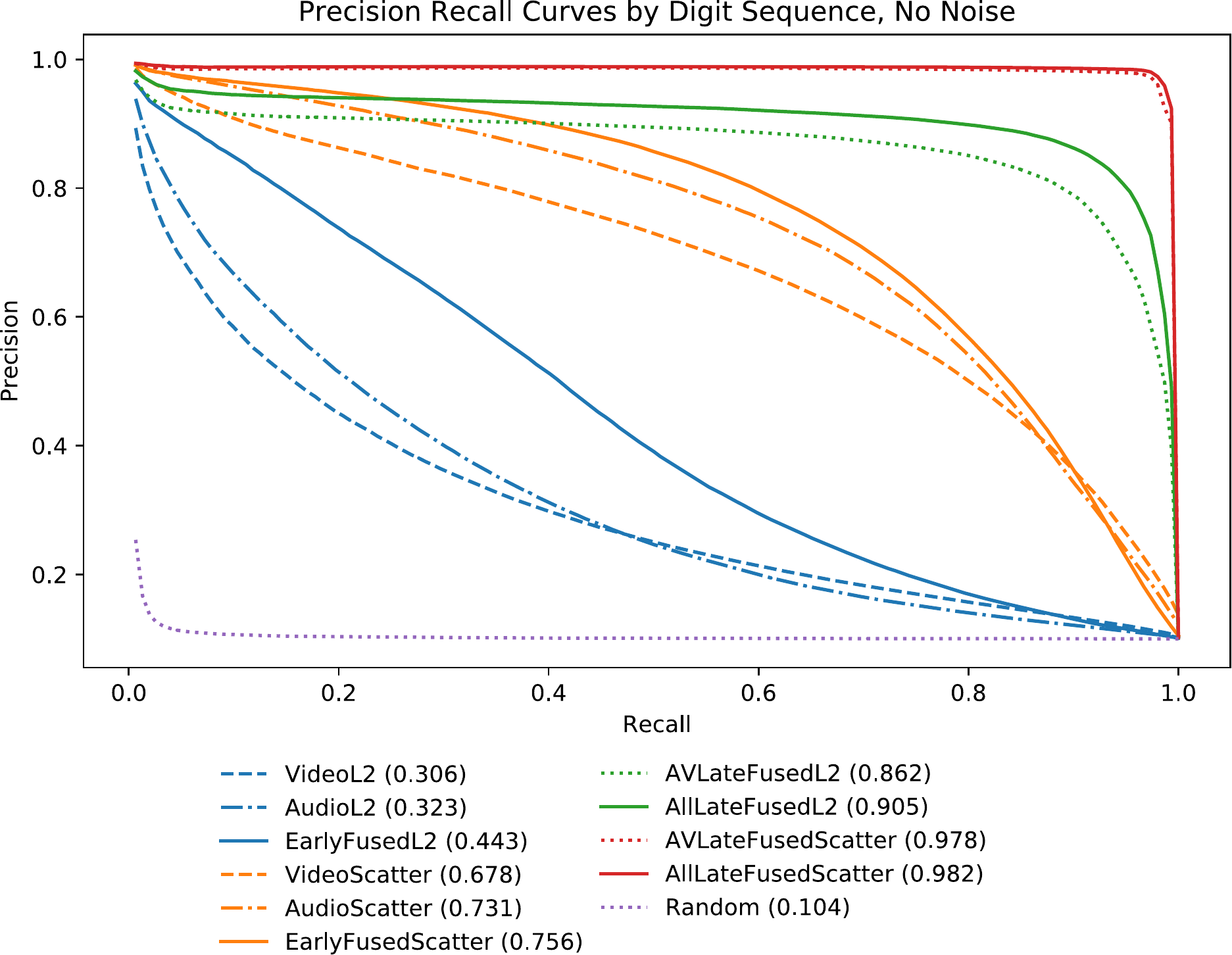}
    \caption{Precision-Recall curves for each pipeline under zero noise. Mean Average Precision (MAP) shown in parentheses.}
    \label{fig:PRresults}
\end{figure}

\begin{figure}
    \centering
    \includegraphics[width=\columnwidth]{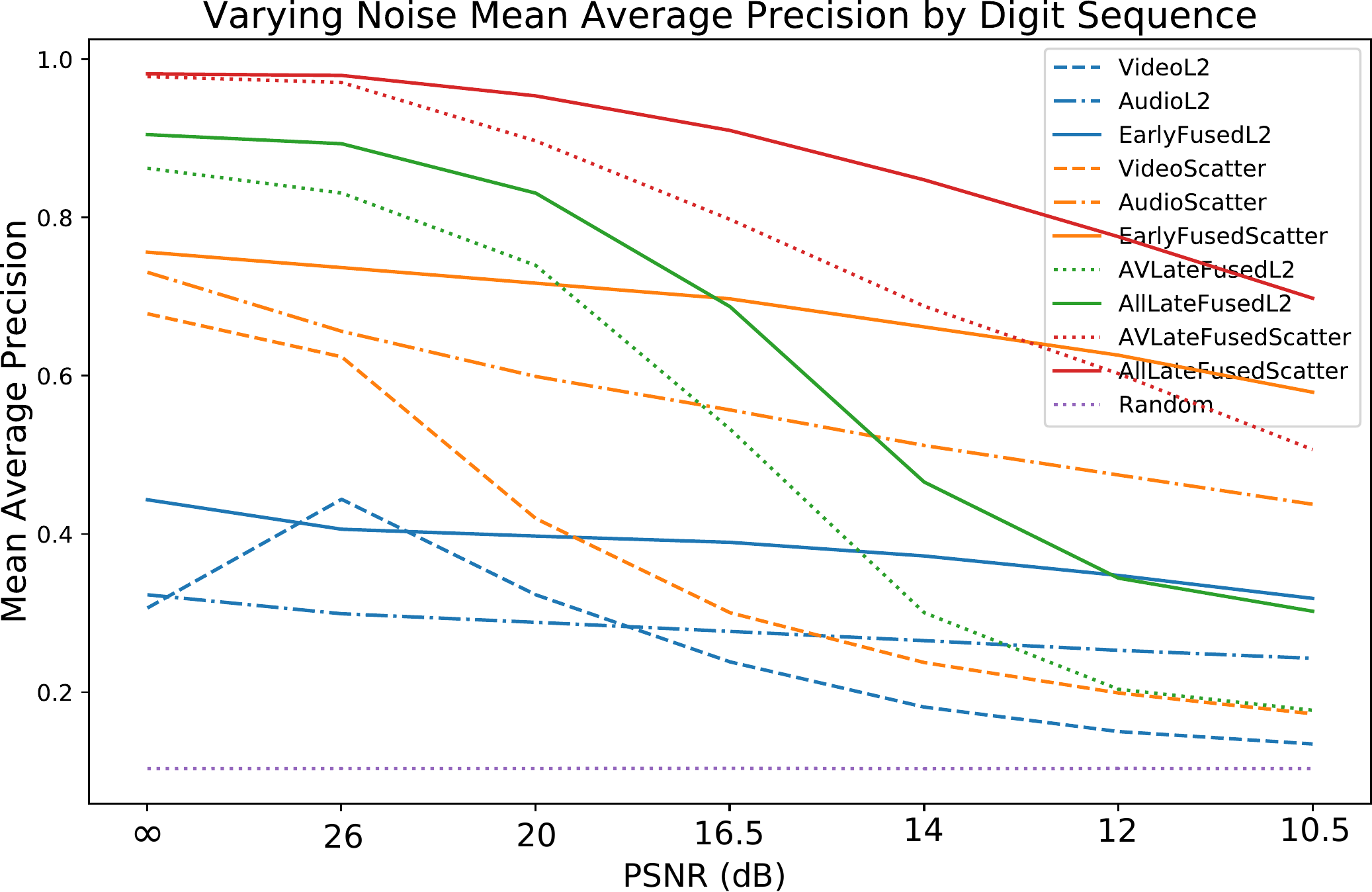}
    \caption{The impact of noise on our pipelines when classifying by digit sequence (regardless of speaker). For each pipeline, mean average precision is plotted against level of noise.}
    \label{fig:NoiseImpactByDigitSequence}
\end{figure}

Precision-recall curves for each pipeline are shown in Figure \ref{fig:PRresults}.
The benefits of using the scattering transform rather than operating directly on the SSMs is clear.
The advantage of fusion before scattering, while less striking, is also apparent.
To probe this further, we ran a second series of experiments, where simulated noise was added to both the raw video and raw audio data before pre-processing.
Figure \ref{fig:NoiseImpactByDigitSequence} shows the results for digit classification plotted against the peak signal to noise ratio (pSNR) in dB ($\infty$ is no noise).  Recall that the pSNR of a signal $x$ is defined as $(20 \log_{10} \max |x|) / \sqrt{MSE(x)}$.  At each level of noise, we compute a P-R curve, but only the MAP values are shown.  The advantage of fusion before scattering increases in the presence of noise.  Furthermore, downstream SNF yields near perfect performance at low levels of noise, and continues to dominate all other pipelines over all noise levels.  

\begin{figure}
    \centering
    \includegraphics[width=\columnwidth]{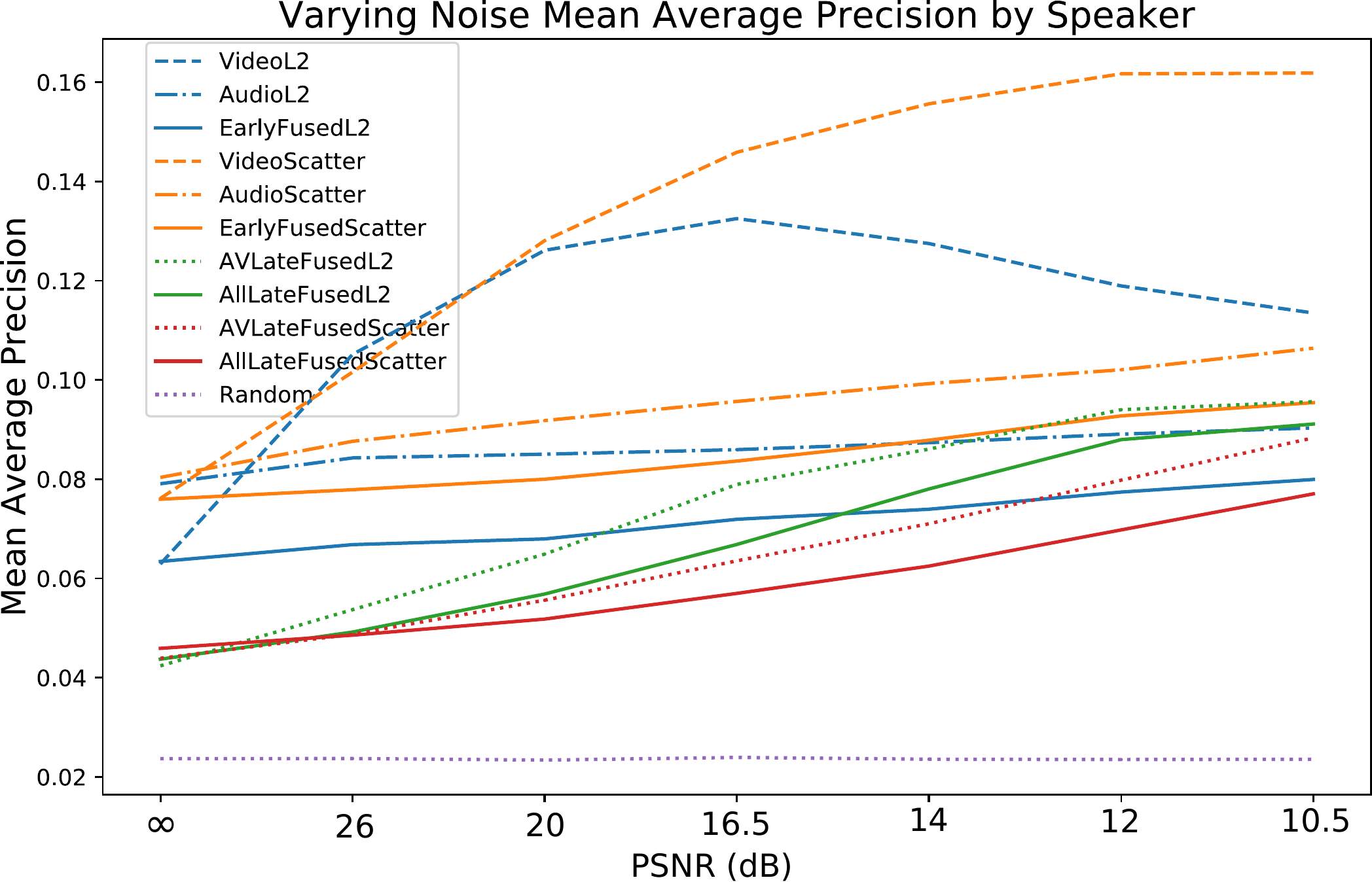}
    \caption{The impact of noise on our pipelines when classifying by speaker (ignoring the digit sequence)}
    \label{fig:NoiseImpactBySpeaker}
\end{figure}

\begin{figure}
    \centering
    \includegraphics[width=\columnwidth]{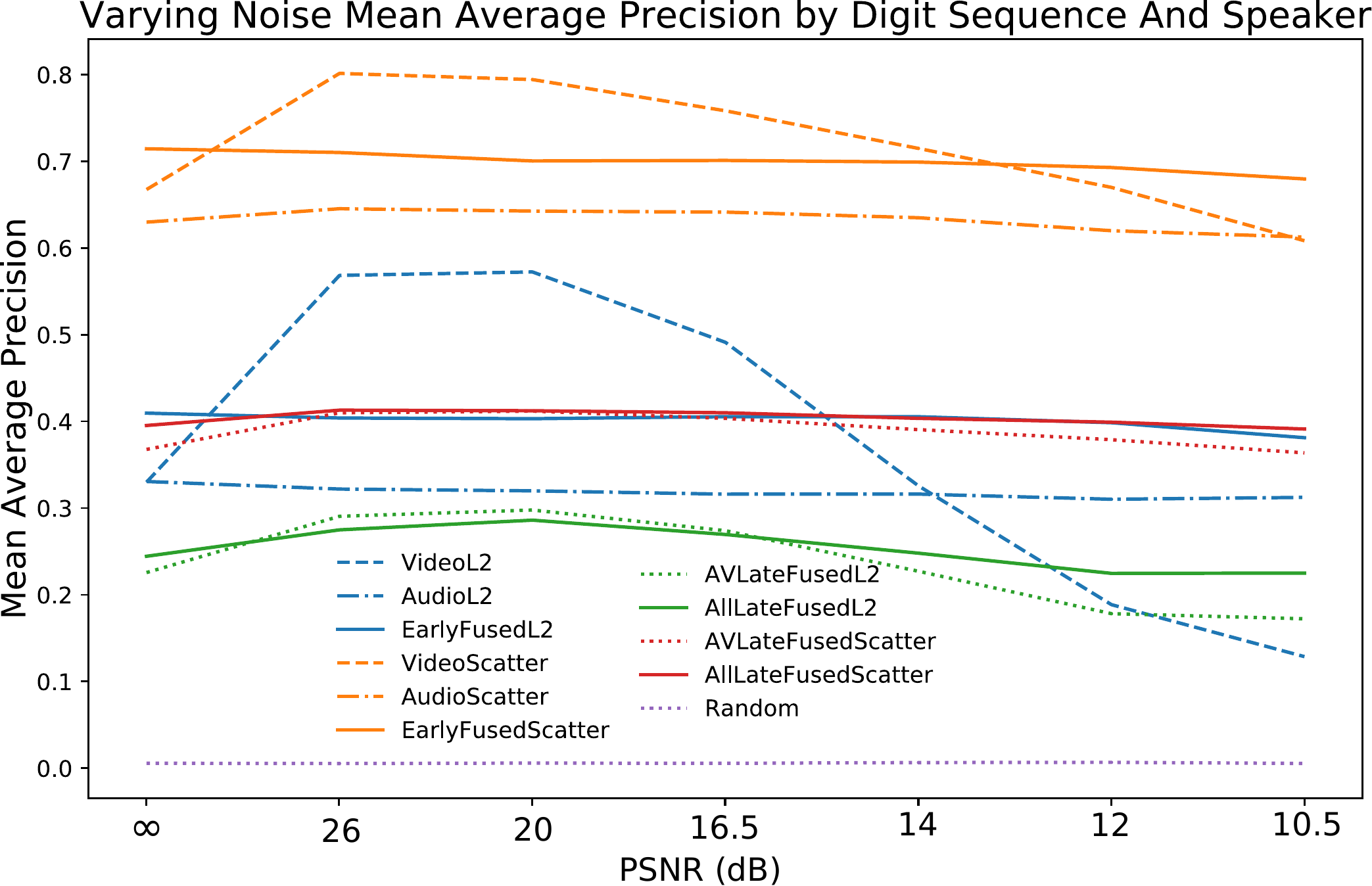}
    \caption{The impact of noise on our pipelines when classifying by speaker {\em and} by digit sequence that they uttered.}
    \label{fig:NoiseImpactByDigitSequenceAndSpeaker}
\end{figure}

Finally, we repeat our noise experiment under different classification schemes to get an even better idea of what SSMs retain.  Based on our experience with cover song analysis \cite{tralie2015coversongstimbral}, we originally hypothesized that SSMs would be approximately invariant to speaker and hence do a particularly poor job of distinguishing speakers
from one another. We do indeed get poor results when attempting to classify which of the 51 speakers is uttering a particular digit, as shown in Figure~\ref{fig:NoiseImpactBySpeaker}.  However, when we drill down even further and attempt to classify which speaker {\em and} which digit sequence is being uttered, we do surprisingly well, as shown in Figure~\ref{fig:NoiseImpactByDigitSequenceAndSpeaker}.  The high MAP values there are particularly striking, as there are 510 classes between which to disambiguate, and there are only 3 examples per class; random guessing only has a MAP of about 0.002.  Figure~\ref{fig:NoiseImpactByDigitSequenceAndSpeaker} does, however, highlight one potential pitfall of downstream SNF, which actually degrades results in all cases.  We believe this is because there are so few examples per class, that the random walk smooths out these finer details.  Hence, we recommend downstream SNF only when one has access to a rich set of examples, with enough examples in each class.

\section{Conclusions}
\label{sec:Conc}

This paper discussed three geometric techniques--Self-Similarity Matrices, Similarity Network Fusion, and the Scattering Transform--and used them together for the first time (to our knowledge) in the fusion of multi-modal time series.
As described above, self-similarity matrices (SSMs) permit the comparison of disparate time series on a common footing, similarity network fusion uses graph-diffusion methods to produce a single SSM that combines the best features of two or more modality-specific SSMs, and the scattering transform extracts multi-scale features from SSMs in a provably robust way.

We constructed several pipelines, involving both upstream feature-level and downstream ranking-level fusion, and demonstrated their benefits on a well-known dataset.
The pipeline performance on digit-string recognition was particularly striking, especially given the unsupervised nature of our methodology.
The specific tests run in this paper used pre-processed streams of video and audio data, but the pipelines are entirely agnostic to choices of modalities and preprocessing thereof.
An important next step is to exploit these methods in other multi-modal contexts, for example to seismic-acoustic fusion \cite{Blasch2011seismic}.

The time complexity of our pipeline is not that bad, as commented on at the end of Sections \ref{sec:FO} and \ref{sec:ST}.
However, this only refers to one possible cost of a fusion pipeline within a notional sensor network.
Assuming a model where there are individual audio/visual sensors and a fusion center, the SNF computation must be done by the fusion center, and modality-specific SSMs must be transmitted to this center from the local sensors.
If transmission is ``expensive'' (e.g,, there may be bandwidth limitation imposed by a mission or there may be detection risks associated with excessive communication), 
then sending entire SSMs may not be advisable.
Some of our ongoing research efforts attacks this problem, by exploring the construction of effective image compression techniques before transmitting the SSMs.

\newpage

\bibliographystyle{plain}
\bibliography{GDAreferences}

\end{document}